\documentclass[conference]{IEEEtran}
\usepackage{graphicx}
\usepackage{float}
\usepackage{cuted}
\usepackage{acronym}

\acrodef{SOTA}{State-Of-The-Art}
\acrodef{SI-SNR}{Scale-Invariant Signal-to-Noise Ratio}
\acrodef{SNR}{Signal-to-Noise Ratio}
\acrodef{GLU}{Gated Linear Unit}
\acrodef{FER}{Facial Emotion Recognition}
\acrodef{LBP}{Local Binary Patterns}
\acrodef{ML}{Machine Learning}

\usepackage{booktabs} 
\usepackage{graphicx}
\usepackage{subfig}

\usepackage[dvipsnames]{xcolor}

\usepackage{float}
\usepackage{wrapfig}

\begin{document}
%
\title{A Peek at Peak Emotion Recognition}

\author{
Tzvi Michelson\textsuperscript{1} \ \ \ \ 
Hillel Aviezer\textsuperscript{2} \ \ \ \  
Shmuel Peleg\textsuperscript{1} \\
\textsuperscript{1}School of Computer Science and Engineering \ \ \ \  
\textsuperscript{2}Department of Psychology\\
The Hebrew University of Jerusalem\\
Jerusalem, Israel
}




\makeatletter
\g@addto@macro\@maketitle{
  \setcounter{figure}{0}
  \begin{figure}[H]
  \setlength{\linewidth}{\textwidth}
  \setlength{\hsize}{\textwidth}
  \centering
  \captionsetup{width=0.96\linewidth}
  \includegraphics[width=1.01\textwidth]{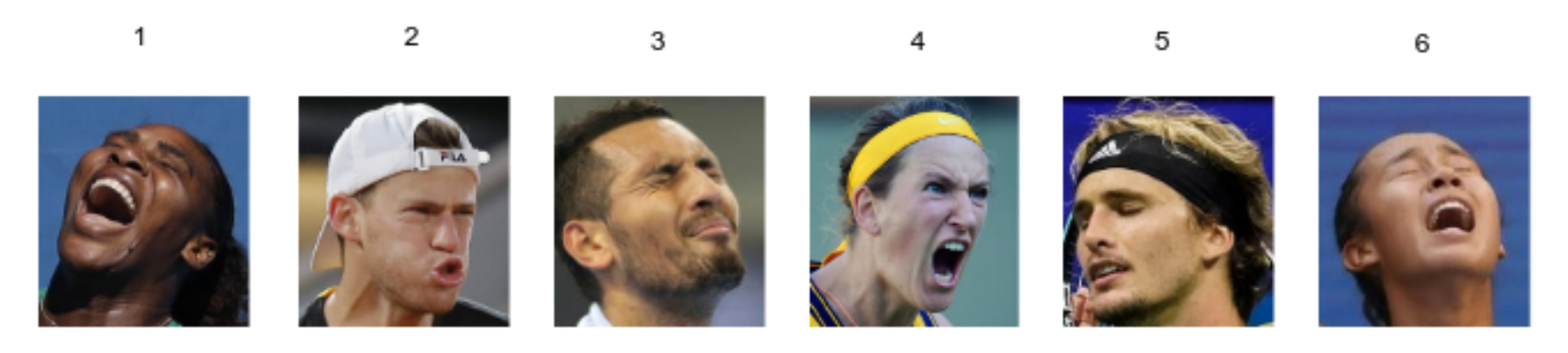}
  \caption{Can you tell in which images the player won a point and in which he / she lost a point? \\ 1, 3, 5 = losing point; 2, 4, 6 = winning point.}
  \label{fig:faces}
  \end{figure}
}
\makeatother

\maketitle

\begin{abstract}
Despite much progress in the field of facial expression recognition, little  attention has been paid to the recognition of peak emotion. Aviezer et al. \cite{aviezer2012body} showed that humans have trouble discerning between positive and negative peak emotions. In this work we analyze how deep learning fares on this challenge. We find that (i) despite using very small datasets, features extracted from deep learning models can achieve results significantly better than humans. (ii) We find that deep learning models, even when trained only on datasets tagged by humans, still outperform humans in this task.
\end{abstract}


%
\IEEEpeerreviewmaketitle

\section{Introduction}
Aviezer et al. \cite{aviezer2012body} discovered that the human ability to distinguish between positive and negative facial expressions in day-to-day situations, does not generalize well to peak (extreme) emotions, such as sports players winning or losing a point. When the emotions are extreme (peak emotions), positive and negative emotions share characteristics such that the distinction becomes significantly harder. In their work \cite{aviezer2012body}, they gather pictures of tennis players winning or losing a point and show the pictures to human evaluators. In one of their settings they show images of only the face, with the body cropped out. In another they show images of only the body, with the face cropped out. In the third setting they show the images as they were taken. They discovered that human raters succeed in distinguishing between winning and losing a point only when they can see the body of the player. However, the peak facial expressions the players display are harder to read. This finding is supported by brain imaging work which finds that the same areas in the brain such as: the insula, striatum, orbitofrontal cortex, and amygdala are activated when experiencing both positive and negative emotions \cite{leknes2008common, said2009nonlinear, adolphs1994impaired, canli2002amygdala}. In this work we study whether neural networks can succeed in correctly classifying the images of the tennis players' faces, despite humans failing. We find that with very little training, features extracted from neural network models are able to predict the classification of the face at a significantly higher accuracy then humans. 


This work is different from most works in the field of \ac{FER} in a number of ways:

\begin{itemize}
    \item First of all, most efforts in the field of \ac{FER} have relied upon datasets with human annotations \cite{li2020deep}. The common datasets include either a categorical classification to a specific emotion (angry, sad, happy etc.) or a score on 2 or 3 dimensions: valence, arousal and sometimes domination. The weakness of these datasets is that they rely on human annotations and are thus limited by them. Human annotations are expensive, can often be imprecise and are only able to detect what humans can detect. Utilizing images of tennis players, immediately after the player either won or lost a point, we can establish an objective, unbiased ground truth.
    \item Second, following Aviezer et al. \cite{aviezer2012body} we continue our evaluation with body pose analysis. Despite raters achieving very high accuracy we do not succeed in training models to the same level. This may be due to the fact that despite much research in the field of \ac{FER}, the field of Pose and Gesture Emotion Recognition is less developed and does not yet have strong models which create good representations. However, it is interesting to note that classifiers trained only on the pose of the hands, surpassed classifiers trained on full body keypoints (excluding hands). The full analysis can be found in Sec.~\ref{sec:body_pose}. Even if this result is because we do not yet have good enough representations of the body, it still shows that a large amount of the information regarding peak emotions can be found in the hands.
    \item Third, by using the results of \ac{FER} models we are to a certain extent comparing human learning and Neural Network learning. If networks trained on human annotation can succeed in this task despite humans failing, it is clear that the function learned by networks to classify a facial expression to its appropriate emotion is different from the function used by humans to perform the same task. In our work we find that even when directly taking predictions from a model trained only on a human annotated data-set, the model can still outperform the human.
\end{itemize}

\begin{table*}[tb]
\centering
\small
\begin{tabular}{lccc}
\toprule
\textbf{Inputs} & \textbf{Model backbone} & \textbf{Dataset for Pretraining} & \textbf{ROC AUC} \\
\midrule
Face & Human Evaluation & & 0.52 \\
\midrule
\textbf{FR models - Features} \\
\midrule
Face & VGG & VGGFace2 & 0.64 \\
Face & Resnet & VGGFace2 & 0.61 \\
Face & SENet & VGGFace2 & \textbf{0.70} \\
\midrule
\textbf{FER models - Predictions} \\
\midrule
Face & VGG & FER+ & \textbf{0.70} \\
Face & Resnet & Affectnet7 & 0.60 \\
Face & Resnet & Affectnet8 & 0.55 \\
Face & Resnet & rafdb & 0.57 \\
\midrule
\textbf{FER models - Features} \\
\midrule
Face & VGG & FER+ & 0.71 \\
Face &  Resnet & Affectnet7 & \textbf{0.72} \\
Face &  Resnet & Affectnet8 & 0.66 \\
Face &  Resnet & rafdb & 0.67 \\
\midrule
\textbf{FER + FR - Features}\\
\midrule
Face & VGG (FER) + VGG (FR) & FER+ / VGGFace2 & 0.71\\
Face & VGG (FER) + Resnet (FR) & FER+ / VGGFace2 & 0.71\\
Face & VGG (FER) + SENet (FR) & FER+ / VGGFace2 & 0.76\\
Face & Resnet (FER) + VGG (FR) & AffectNet7 / VGGFace2 & 0.72\\
Face & Resnet (FER) + Resnet (FR) & AffectNet7 / VGGFace2 & 0.72\\
Face & Resnet (FER) + SENet (FR) & AffectNet7 / VGGFace2 & \textbf{0.78}\\
\bottomrule
\end{tabular}
\vspace{0.25cm}
\caption{FR = Facial Recognition, FER = Facial Emotion Recognition \\
For every model, the table describes it's architecture, the data it was pretrained on and the ROC AUC achieved on our data, using 5-fold cross validation. In the case of the FER models we examined the options of (i) using the output of the models directly i.e. happy is winning and sad is losing. This option is called "FER Model - Predictions". (ii) Extracting features on which to train a classifier, this option is called "FER Models - Features".}
\label{tab:face}
\end{table*}

\section{Related Work}

\subsection{Automatic Facial Expression Recognition}
The field of automatic \ac{FER} can be generally split into 2 main categories: discrete and continuous. The first people to identify discrete, universal emotions were Ekman and Friesen \cite{ekman1971constants} who identified 5 discrete emotion: happiness, sadness, anger, surprise, disgust and fear. They claimed these emotions had universal facial expression, consistent across cultures. Contempt was later added as another emotion with universal facial expressions \cite{matsumoto1992more}. Although recent research has cast doubt on the universality of the the expressions \cite{jack2012facial}, most works still use this framework for the \ac{FER} task. A number of datasets have been gathered and manually annotated with these emotions. Examples of such datasets include:
\vspace{0.02in}
\begin{itemize}
    \item \textbf{CK+} \cite{lucey2010extended} - contains multiple series of images in which the face changes from neutral to expressive \\
    \item \textbf{FER2013} \cite{goodfellow2013challenges} - contains 35,000 individual images each tagged with one of seven emotions \\
    \item \textbf{AffectNet} \cite{mollahosseini2017affectnet} - contains over a million images obtained by querying search engines for emotion-related tags, 450,000 of the images have manually annotated labels for eight basic emotions
\end{itemize}
and other less well known datasets.

The second category is continuous affect recognition. In this framework facial expressions are placed on a 2-dimensional grid where one axis represents {\bf arousal} (relaxed vs. aroused) and the other {\bf valence} (pleasant vs. unpleasant) \cite{gunes2013categorical}. Datasets annotated according to this model include: \textbf{SEWA} \cite{kossaifi2019sewa} which contains over 2000 minutes of hand annotated audio-visual data and \textbf{Facial Affect “in-the-wild”} \cite{zafeiriou2016facial} which gathered data from 500 youtube videos and manually annotated them with regard to valence and arousal.

Both of these categories were originally analyzed using classical methods such as: \ac{LBP} \cite{shan2009facial}, \ac{LBP} on three orthogonal planes \cite{zhao2007dynamic}, non-negative matrix factorization (NMF) \cite{zhi2010graph} and sparse learning \cite{zhong2012learning}. Since 2013 emotion  recognition competitions have collected sufficient training data from challenging real-world scenarios,  which implicitly promote the transition of FER from lab-controlled to in-the-wild settings. Additionally, deep learning methods provided the tools to utilize these datasets and have achieved state-of-the-art results \cite{kaya2017video, knyazev2017convolutional}. Their methods involve using a model pretrained on VGGFace or Imagenet and then fine tuned with FER specific datasets. The backbone model is usually convolutional and is heavily influenced by state of the art Facial Recognition models \cite{li2020deep}.

\subsection{Pose and Gesture Emotion Recognition}
Pose and Gesture Emotion Recognition is significantly less studied then \ac{FER} \cite{noroozi2018survey}. Here, in a similar fashion to \ac{FER}, emotions are either defined categorically where the goal is to classify the pose to the correct emotion, or continuously, where every pose is given a score on a few continuous scales. Unlike \ac{FER}, very little work has been done to use deep learning for Pose and Gesture Emotion Recognition\cite{noroozi2018survey}. Also, the datasets here are less standardized, and most works use data they gathered themselves.

Using classical methods, Saha et al. \cite{saha2014study} investigated gestures reflecting five basic human emotional states from skeletal geometrical features. They compared a variety of machine learning classifiers and obtained best results using ensemble trees. They gathered their data by stimulating the desired emotion in their study participants while recording the motions and gestures with a Microsoft Kinect (depth sensor). Fourati and Pelachaud \cite{fourati2015multi} proposed a different approach to analyse the emotional meaning of movements by using a range of features from the whole body. They described movement on an anatomical, directional and posture level. Using these descriptors, they used a Random Forest classifier to classify the movements to one of eight emotional states expressed by actors in various day-to-day actions such as walking, sitting etc.
Moving on to deep methods, Kosti et al. \cite{kosti2017emotion} presented a method for emotion recognition based on images containing people in non-controlled environments. They trained a two low-rank filter CNN that jointly analysed the person and the whole scene to recognize the emotional state. The analysed images depicted people annotated with 26 emotional categories as well as the continuous dimensions: valence, arousal, and dominance. In their research they emphasized the importance of context for recognizing people’s emotions in images.
A few works have used Body Pose in multimodal affect recognition together with \ac{FER} \cite{gunes2007bi} or audio-based affect recognition \cite{vu2011emotion}. 

\section{Data}
The data used in this paper consists of 176 images of players in tennis matches. 88 of the images depict a player winning a point and 88 depict a player losing a point. The images were obtained using a google image search and the queries "reacting to a winning point" or "reacting to a losing point", crossed with "tennis". The gender distribution for the winning images is 41 male vs 43 female, and for the losing images is 45 male vs 39 female. Although some of the tennis players were more popular then others there are 71 unique tennis players losing a point and 75 unique tennis players winning a point. Removing the tennis players appearing more then once did not significantly affect the results of the analysis. 

In the original analysis by Aviezer et al. \cite{aviezer2012body} the faces were cropped by hand, in our work we used an automatic face detector \cite{zhang2016joint} since deep learning models require a rectangular image to function properly. Images chosen by journalists to depict losing or winning a point typically contain the image representing the peak emotion \cite{aviezer2012body}. This has been confirmed with a manual analysis of randomly sampled professional tennis matches from the finals of Wimbledon and Australia open (full matches were captured from YouTube) \cite{aviezer2012body}. The focus was put on the most critical emotional peak of the events: the immediate response to winning the final match points. The faces of the winners/losers were essentially neutral immediately before the final serve of the game. However, upon winning/losing the match, a spike in facial movement was observed. The facial expressions of the winners replicated the type of reactions found in the image search. Importantly, these intense expressions reached their peak less than 1000ms after the onset of the winning/losing. The rapid elicitation of the expression immediately after the emotional event confirms that these expressions indeed reflect a peak experience.

It was found \cite{aviezer2012body} that when examining images in this dataset humans had a very difficult time distinguishing between faces of winning and losing players, in cases where only the faces were shown to them (ROC Auc 0.52). At the same time, when only the body of the player was shown to the participants they achieved significantly better results (ROC Auc 0.93). In the next section we will describe our proposed methods to automatically analyze the players' expressions.

\begin{figure*}[!t]
\centering
\includegraphics[width=7in]{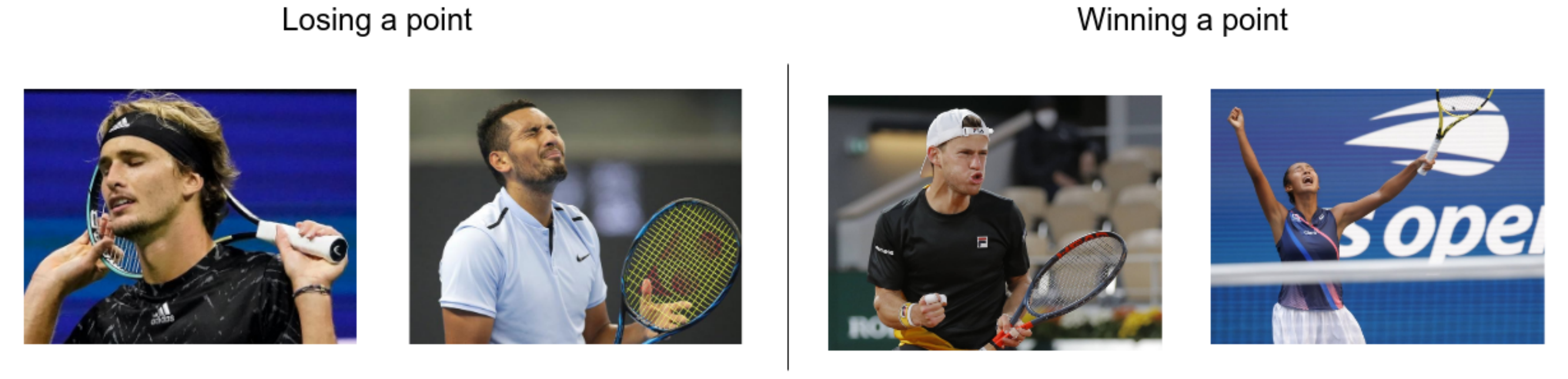}
\hfil
\caption{Images of the tennis players including body pose. The cropped faces can be found in figure~\ref{fig:faces}. In this scenario it is significantly easier to discern which images are of losing players and which are of winning players.}
\label{fig:body}
\end{figure*}

\section{Modality I - Face}
\label{sec:face}

Due to lack of data it was impossible to train a model from scratch, so we made extensive use of pretrained models. In all cases features were extracted from the pretrained model and a simple \ac{ML} classifier was trained on the extracted features with heavy regularization. Also, in all cases which involved training the evaluation was done using 5-fold cross validation to calculate the ROC Auc. Previous research has shown that embedded in the faces there are clues to the quality of the emotion \cite{aviezer2015thrill}, here we attempt to see how extensive these clues are and how well neural networks can utilize them.
\subsection{FR - Face Recognition Models}
In this stage we used 3 different face recognition models each trained on VGGFace2 \cite{cao2018vggface2}. The VGGFace2 dataset contains 3.31 million images of 9131 subjects, with an average of 362.6 images per subject. Images were downloaded from Google Image Search and have large variations in pose, age, illumination, ethnicity and profession (e.g. actors, athletes, politicians etc.). 
The models we used are a VGG16 \cite{simonyan2014very} from which we extracted a 512-d feature vector, a Resnet50 \cite{he2016deep} from which we extracted a 2048-d feature vector and a SEnet50 \cite{hu2018squeeze} from which we extracted a 2048-d feature vector. The SENet50 is a variation of the Resnet which includes squeeze-excite blocks. These blocks allow global information to impact the learning after every convolutional block. After experimenting with a number of classifiers we trained a Logistic Regression classifier on the extracted vectors and used 5-fold cross validation to calculate ROC AUC. The results can be found in Table~\ref{tab:face}. We used face recognition models since they have been shown to extract good representations of the face \cite{ephrat2018looking} and have been used in a variety of other tasks which require such representations \cite{michelson2021audio, ephrat2018looking}.

\subsection{FER - Facial Emotion Recognition Models}
Here, we experimented with networks which were essentially trained for the precise task we are trying to perform - \ac{FER}. We use a VGG13 network trained by Barsoum et al.~\cite{barsoum2016training} and a Resnet18 network trained by Wen et al.~\cite{wen2021distract}. The networks are trained on 4 different datasets: FER+~\cite{BarsoumICMI2016}, Affectnet7~\cite{mollahosseini2017affectnet}, Affectnet8~\cite{mollahosseini2017affectnet} and rafdb~\cite{li2017reliable}. We performed our experiments in 2 parts. 
\begin{enumerate}
    \item In the first part we took the output of the network before the softmax, for the emotions happy and sad and inputted just the two scores into the softmax function. Thus, we could essentially see what are the raw predictions of the network for this face. This part achieved results better then human performance when using the FER+ dataset~\cite{BarsoumICMI2016}. This is of interest since the dataset FER+~\cite{barsoum2016training} was tagged by humans, meaning the network is essentially learning to approximate the human evaluation function, and yet still when predicting on peak emotion the network outperforms humans.
    \item In the second part we extracted a feature vector from the network and trained a classifier on the feature vector, we experimented with a number of classifiers and found that usually Logistic Regression works best.
\end{enumerate}
A summary of the results can be found in Table~\ref{tab:face}.

\begin{table*}[tb]
\centering
\small

\begin{tabular}{lccc}
\toprule
\textbf{Inputs} & \textbf{Model backbone} & \textbf{ROC AUC} \\
\midrule
Body & Human Evaluation & 0.93 \\
Body + Face & Human Evaluation & 0.91 \\
\midrule
\textbf{Body pose}\\
\midrule
Body Pose (3D) & Open Pose & 0.60\\
Body Pose (3D) & Detectron2 &  \textbf{0.75}\\
\midrule
\textbf{FER + FR + Body pose}\\
\midrule
Face + Body Pose & Face + Open Pose & 0.80\\
Face + Body Pose & Face + Detectron2 & \textbf{0.83}\\
\midrule
\textbf{Hand pose}\\
\midrule
Hand Pose & MediaPipe & 0.83\\
Hand Pose + Body Pose & MediaPipe + Detectron2 & 0.83\\
Hand Pose + Face & MediaPipe + Face & \textbf{0.85}\\
Hand Pose + Face + Body Pose & MediaPipe + Detectron2 + Face & \textbf{0.85}\\
\bottomrule
\end{tabular}
\vspace{0.25cm}
\caption{This table includes the ROC Auc scores achieved by classifiers trained on body and hand pose keypoints extracted from the images, as well as the different combinations. We can see that the hand keypoints contain almost all of the pose information. When face is mentioned in this table, it referrences the best performing model from Sec~\ref{sec:face}.}
\label{tab:body}
\end{table*}

\subsection{Implementation Details}
In all cases where the features were extracted from a network they were then fed into a classical \ac{ML} classifier. We experimented with the following classifiers: Random Forest, Logistic regression, K-nearest neighbors and SVM. A randomized grid search was run over the following hyperparameters: Regularization type (L2, L1 or elasticnet, for Logistic Regression), Regularization Coefficient, Kernel (polynomial or linear, for SVM), tree depth (for random forest) and number of neighbors (for K-nearest neighbors). The same methodology was used for body pose as well.

\section{Modality II - Body pose}
\label{sec:body_pose}

\begin{figure}[tbh]
\centering
\includegraphics[width=3.1in]{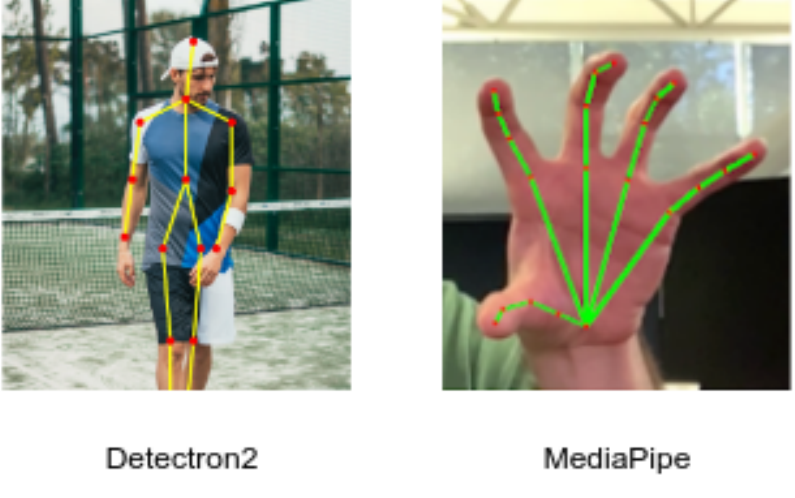}
\hfil
\caption{Visualizations of full body pose coordinates vs. only hand coordinates.}
\label{fig:hand}
\end{figure}

\subsection{Full Body Pose}
An additional input that was studied  by Aviezer et al. \cite{aviezer2012body} is the body pose. In their experiments subjects were able to distinguish between winners and losers using body pose alone. Additionally, they found that the combination of facial expressions with body pose deteriorated performance slightly. In our experiments we found, that models are less successful then humans. We used 2 different pre-trained models \cite{wei2016cpm, 8765346, wu2019detectron2} which extracted 3D coordinates for the main body keypoints and then we trained a classifier on these keypoints to predict whether they came from a winning player or a losing player. The first model we used is OpenPose \cite{wei2016cpm, 8765346}. OpenPose works in two steps. In step 1 it identifies different body points within the image, as well as the fields connecting the body points. In the second step it follows a greedy inference process for connecting the keypoints through said fields. The second model we use, Detectron2, uses He et al.'s mask RCNN \cite{he2017mask} to output a one hot image size filter for every keypoint of every person detected in the image, and uses a whole human detector to connect the keypoints. The 2D coordinates outputted by Detectron2 are then fed to another network which parses them and coverts them to 3D coordinates. A K-nearest neighbors classifier worked best for these features. Our results can be found in Table~\ref{tab:body}.

\begin{figure*}[!t]
\centering
\includegraphics[width=6.5in]{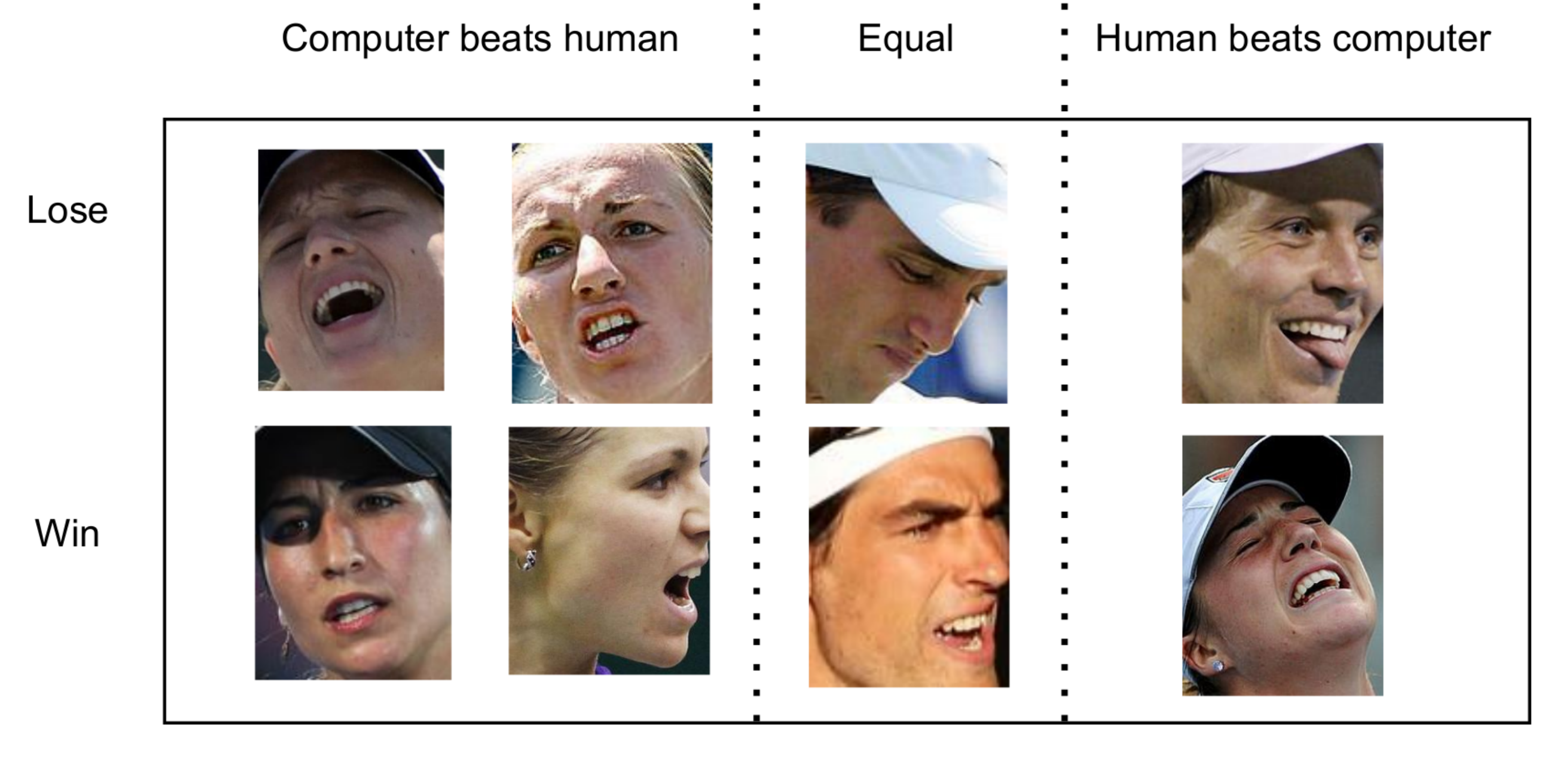}
\hfil
\caption{Examples of images in six different categories. Top row is images of players who just scored (won) a point. Bottom row is images of players who were scored (lost) a point. Left area includes images which our model classifies correctly and humans classify incorrectly. Middle segment includes images where both humans and our model classify correctly. Right area includes images where humans classify correctly and our model classifies incorrectly.}
\label{fig:comparisons}
\end{figure*}

\subsection{Hand Pose}

A surprising weakness of most models which extract body pose is that their analysis is limited to major body parts i.e. arms, legs, head, shoulders and hips while the fingers and hand are ignored. A visualization of this can be seen in the left side of Figure~\ref{fig:hand}. When examining our images we noticed that winning players often fist pumped. This additional information is intuitively utilized by human viewers but can not be used by our model which is unaware of the finger locations. Thus we used Google's MediaPipe \cite{zhang2020mediapipe} framework to extract hand keypoints and then trained a classifier on these keypoints to predict whether the player lost or won a point. Our results show that indeed the hand pose provide more information then the general body pose, and in fact almost all the information which can be extracted from the body pose allready exists in the hand pose. The quantitative results can be seen in Table~\ref{tab:body}.

\section{Combining different modalities}
To estimate the total ability of models in this task we experimented with the combination of different modalities including Facial Recognition + Facial Emotion Recognition + Body Pose + Hand Pose. The combination was generally done by averaging across the best performing results from previous stages as this achieved better results then training a classifier on the concatenated features. As can be seen in Table~\ref{tab:body} we found that most of the information models can capture from the body and hand pose is allready captured from the face. It is noteworthy that in the case of body and hand emotion recognition we were unable to find models trained specifically for this task and as such unable to extract domain-relevant features. This is likely the cause of the gap between the humans' and the models' performance.

\begin{center}
\captionsetup{type=figure}
  \captionsetup{width=.9\linewidth}
  \includegraphics[width=\linewidth]{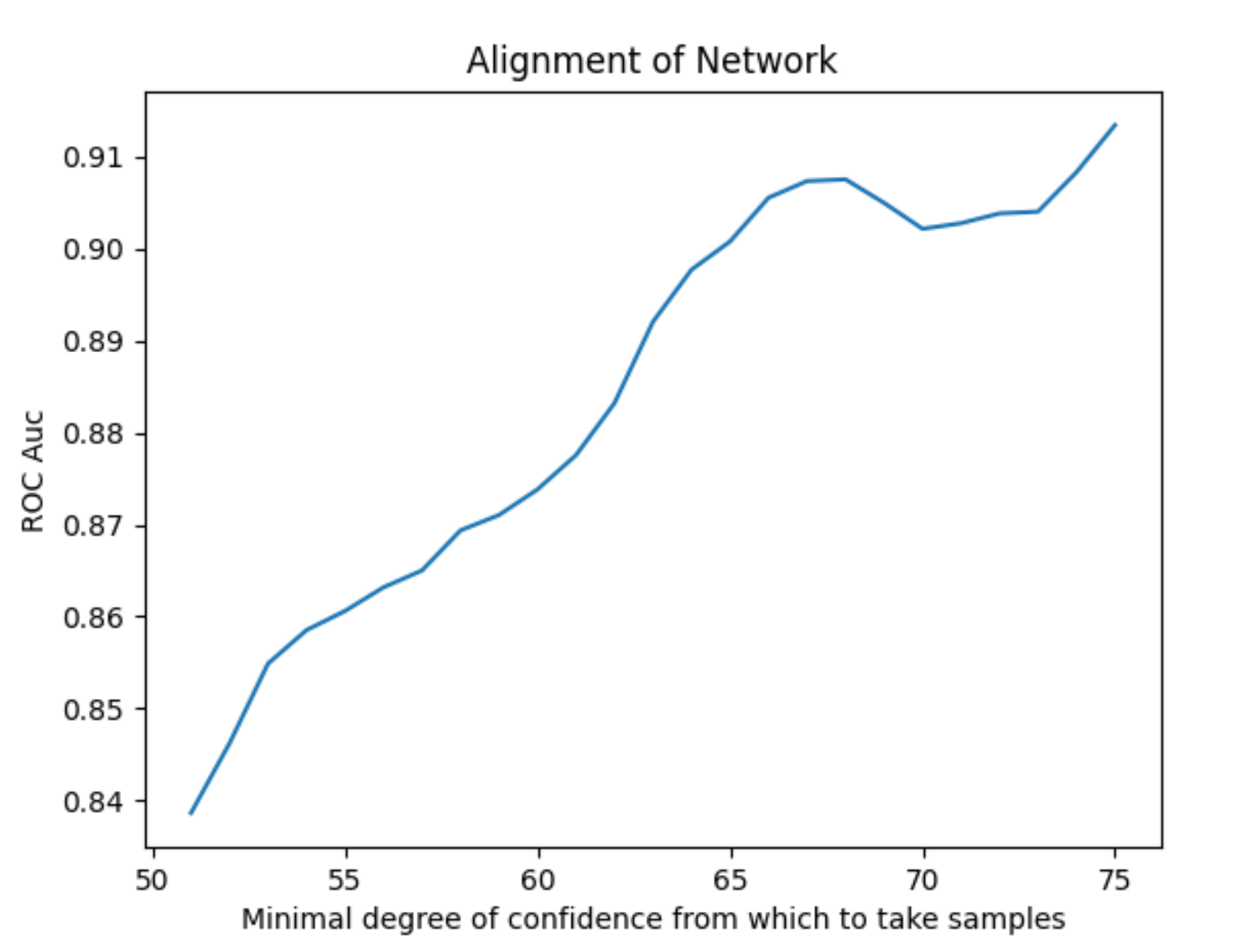}
  \caption{Y axis describes ROC Auc score, X axis describes the certainty score from which samples were accepted. For example: certainty score 50 means all samples, certainty score 75 means only predictions of above 0.75 and below 0.25 are used for the analysis.}
\label{fig:alignment}
\end{center}

\subsection{Alignment}
To further analyze our model we tested it's alignment. What we were trying to understand is whether the model is more correct when it is more confident in it's prediction. To test this, we calculate the ROC Auc scores while only taking the samples where the model's confidence is above a certain threshold. We run $i$ from 50 to 75, and only take the samples with a score greater then i (high confidence in a positive score) or lower the 100 - i (high confidence in a negative score):  \vspace{0.1in} \\
$for\ i\ in\ range(50, 75):$
\[currSamples = \{x \in Samples\ |\ x > i\ or\ x < 100 - i \}\] 
For each iteration we calculate the ROC Auc score and plot them. We do not go above 75\% confidence since beyond that point there are not enough samples for the ROC Auc score to be meaningful. We find our model to be well calibrated, where if you only take predictions where the model is sure of itself, the ROC Auc score increases to 0.91. A graph of these findings can be found in Figure~\ref{fig:alignment}.

\section{Qualitative comparison}

Examining images where our model made errors vs. images on which humans made errors can supply insight into attributes of the function the model used. Figure~\ref{fig:comparisons} provides examples of cases of both losing and winning players where: (1) humans classify incorrectly while our model classifies correctly, (2) our model as well as humans classify correctly and (3) humans classify correctly while our model classifies incorrectly. To get a taste of why the model makes mistakes we will examine the two images in the figure. In the top right corner of the figure we have a case where the player is sticking out his tongue and the model made an error. This image is one of only 3 in the dataset where the player is sticking out his tongue, and sticking out the tongue is generally associated with positive and not negative emotions \cite{mendez2020positive}. As such that it is easy to understand why pretrained networks would mistake this expression for a positive one. The bottom right image presents a female tennis player wearing a hat while looking up with an intense expression on her face. Her mouth is open, her eyes squeezed shut and her brow is furrowed. It is clear that this is an intense expression, however it is less clear if the expression is positive or negative. In this case the player lost a point, the human classified her correctly while our model mistook her for winning a point. However, if you examine the top left image in figure~\ref{fig:comparisons} the description provided above fits her identically. All the way from the female player wearing a hat, until the furrowed brow. Here, our model classifies her correctly as having won a point while human annotators classify her as having lost a point. We were not able to see a qualitative difference between the two images so perhaps there are indeed expressions which are inherently ambivalent.

\section{Conclusion}
We find, that models can perform better then humans on tasks when provided with sufficient relevant training data. We also find that even when models are trained on data tagged by humans, they can still outperform the human taggers, likely by learning patterns which for some reason humans do not. 







%
{
\bibliographystyle{IEEEtran}
\bibliography{refs}
}




\end{document}